%% file: main.tex
\documentclass[10pt,twocolumn,letterpaper]{article}

\usepackage{iccv}              


\definecolor{iccvblue}{rgb}{0.21,0.49,0.74}
\usepackage[pagebackref,breaklinks,colorlinks,allcolors=iccvblue]{hyperref}
\usepackage{algorithm}
\usepackage{algpseudocode}
\usepackage{multirow}
\usepackage{pifont}
\usepackage{xcolor}
\usepackage{adjustbox}
\usepackage{colortbl}
\usepackage{caption}
\usepackage{makecell}
\usepackage{listings}
\usepackage{tabularx}
\definecolor{graybg}{rgb}{0.95, 0.95, 0.95}
\lstdefinestyle{jsonstyle}{
    backgroundcolor=\color{graybg},
    basicstyle=\ttfamily\footnotesize,
    numbers=left,
    numberstyle=\tiny\color{gray},
    stepnumber=1,
    breaklines=true,
    frame=single,
    captionpos=b
}

\def\paperID{9140} 
\def\confName{ICCV}
\def\confYear{2025}

\newcommand{\yingchen}[1]{{\color{red}{\bf\sf [yingchen]}}}
\newcommand{\zuhao}[1]{{\color{yellow}{\bf\sf [zuhao]}}}

\title{TimeExpert: An Expert-Guided Video LLM for Video Temporal Grounding}

\author{
    Zuhao Yang$^{1}$\footnotemark[1]\quad
    Yingchen Yu$^{2}$\quad
    Yunqing Zhao$^{2}$\quad
    Shijian Lu$^{1}$\footnotemark[2]\quad
    Song Bai$^{2}$ \\ [0.5em]
    $^{1}$Nanyang Technological University\qquad
    $^{2}$ByteDance Inc. \\ [0.5em] \href{https://mwxely.github.io/projects/yang2025time/index}{\textcolor{NavyBlue}{https://mwxely.github.io/projects/yang2025time/index}}
}

\begin{document}
\twocolumn[{
    \renewcommand\twocolumn[1][]{#1}
    \maketitle
    \begin{center}
        \centering
        \vspace{-5mm}   
        \captionsetup{type=figure}
        \includegraphics[width=\textwidth]{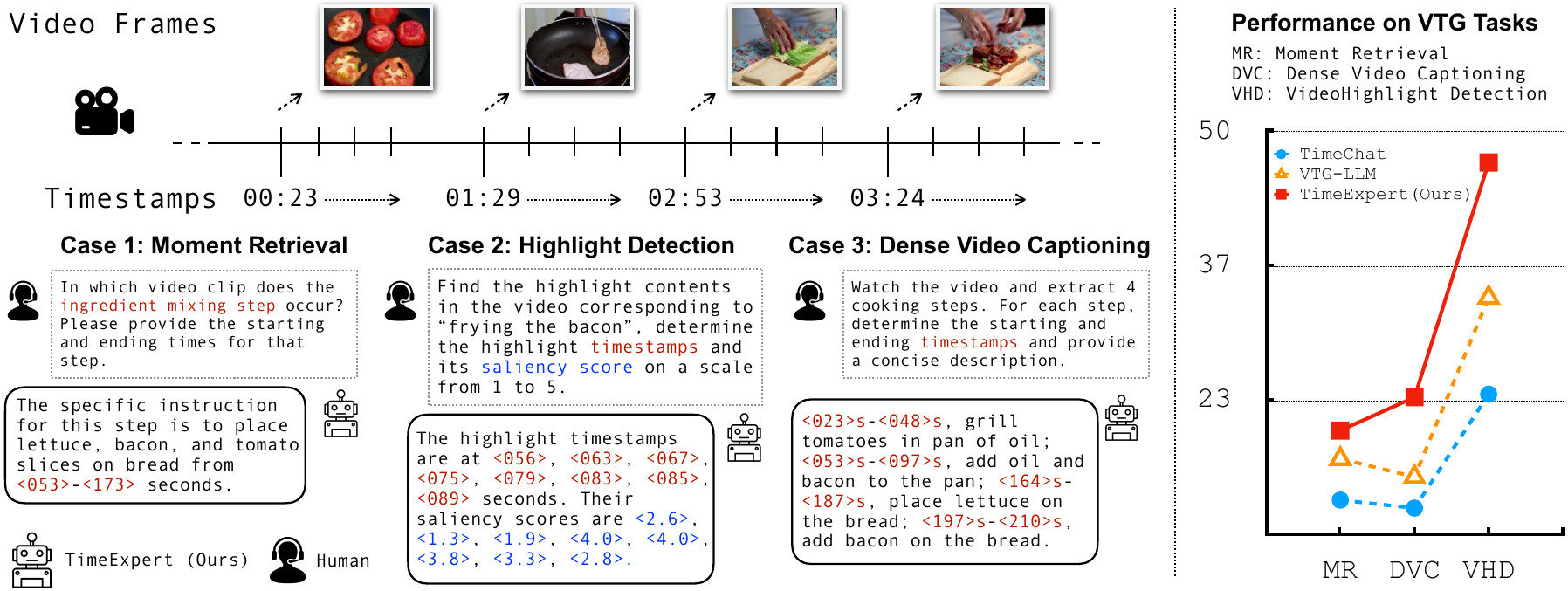}
        \captionof{figure}{
        {\bf Left:} 
        Video Temporal Grounding (VTG) is a fine-grained video understanding task that aims to accurately locate content along with event timestamps based on natural language queries. 
        In this work, we mainly consider three major types of VTG tasks: (1) Moment Retrieval (MR), (2) Video Highlight Detection (VHD), and (3) Dense Video Captioning (DVC). 
        The outputs of VTG often contain textual captions, \textcolor{BrickRed}{timestamps}, and \textcolor{blue!75}{saliency scores}. 
        {\bf Right:} 
        Unlike existing methods (e.g., TimeChat \cite{ren2024timechat}) that employ a single \textit{static} model, motivated by expert specialization on different task tokens \cite{guo2024trace}, we propose \textsl{TimeExpert}, an expert-guided Video LLM with \textit{dynamic} token routing.
        Through task-aware expert allocation, TimeExpert demonstrates substantial improvements over state-of-the-art Video-LLMs on several VTG benchmarks.
        For example, here we visualize zero-shot F1 score for DVC on the YouCook2 dataset \cite{zhou2018towards}, $\text{R@}1_{\text{IoU}=0.7}$ for MR on the Charades-STA dataset \cite{gao2017tall}, and $\text{HIT@}1$ for VHD on the QVHighlights dataset \cite{lei2021detecting}. More results and analysis are in \Cref{sec:exp}.
        }
        \label{fig:teaser}
    \end{center}
}]
{\renewcommand*\thefootnote{\fnsymbol{footnote}}
    \footnotetext[1]{This work was done while Zuhao Yang was interning at ByteDance.}
    \footnotetext[2]{Shijian Lu is the corresponding author.}
}
\input{sec/0_abstract}    
\input{sec/1_intro}
\input{sec/2_related}
\input{sec/3_method}
\input{sec/4_exp}
\input{sec/5_conclusion}
\section*{Acknowledgements}
This study is supported by the Ministry of Education Singapore, under the Tier-1 project scheme with the project number RT18/22.

{
    \small
    \bibliographystyle{ieeenat_fullname}
    \bibliography{main}
}

\end{document}

%% file: sec/0_abstract.tex
\begin{abstract}
Video Temporal Grounding (VTG) aims to precisely identify video event segments in response to textual queries.
The outputs of VTG tasks manifest as sequences of events, each defined by precise timestamps, saliency scores, and textual descriptions.
Despite recent advances, a fundamental limitation persists in existing Video Large Language Models (Video-LLMs): they process all task tokens through identical and static pathways, failing to recognize that temporal localization, saliency assessment, and textual generation represent fundamentally distinct tasks requiring specialized processing.
To address this, we introduce TimeExpert, a Mixture-of-Experts (MoE)-based Video-LLM that effectively decomposes VTG tasks by dynamically routing task-specific tokens (e.g., timestamps, saliency scores) to specialized experts, with increased computational efficiency. 
Our design choices enable precise handling of each subtask, leading to improved event modeling across diverse VTG applications.
Extensive experiments demonstrate that TimeExpert consistently achieves state-of-the-art performance on various VTG tasks such as Dense Video Captioning, Moment Retrieval, and Video Highlight Detection.
\end{abstract}

%% file: sec/1_intro.tex
\begin{figure*}[t]
    \centering
    \includegraphics[width=\textwidth]{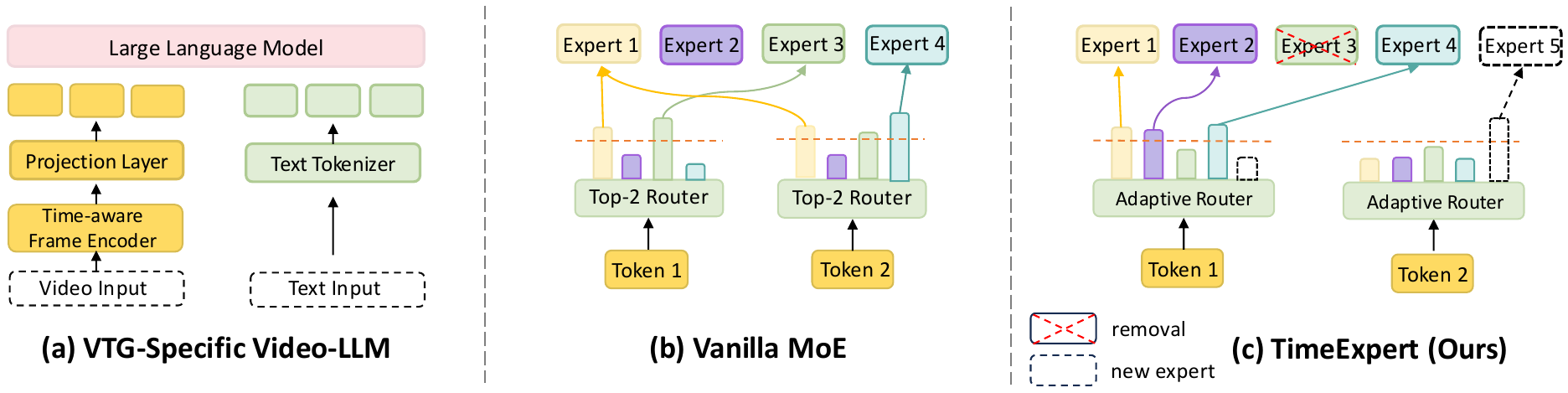}
    \caption{
    {\bf A Glimpse of Comparison across VTG Approaches.}
    {\bf (a):}
    VTG-specific Video-LLM \cite{ren2024timechat,guo2024trace} relies on a single static model with shared parameters for all tasks, limiting their ability to specialize in diverse VTG subtasks. 
    {\bf (b):}
    Vanilla MoE improves upon this by activating a fixed set (e.g., $k$=2) of experts, enabling some degree of task specialization. 
    {\bf (c):}
    Our \textsl{TimeExpert} takes a step further by implementing an adaptive routing mechanism that dynamically allocates new experts when tokens lack suitable matches and prunes unmatched experts when necessary. 
    This dynamic approach significantly enhances computational efficiency while achieving superior task specialization, particularly for VTG subtasks that require distinct feature representations.
    Detailed design choices can be found at \Cref{sec:method}.
    }
    \label{fig:method}
\end{figure*}
\vspace{-1.5em}
\section{Introduction}
\label{sec:intro}

Consider watching a cooking tutorial video and attempting to pinpoint the exact moment when the chef fries the bacon (\Cref{fig:teaser}-Left) or other highlighted visual segments.
This seemingly simple task reveals a fundamental challenge: while humans can effortlessly recognize these actions visually, determining their precise temporal boundaries—down to the specific second—remains significantly challenging even for human observers.
This problem forms the core motivation behind Video Temporal Grounding (VTG), which aims to develop intelligent systems capable of accurately locating relevant video segments based on natural language queries, enabling users to efficiently access desired content without manually searching through hours of footage.

\noindent{\bf Research Gaps.}
While traditional video grounding models excel in individual VTG subtasks like Moment Retrieval~\cite{han2024unleash,lei2021detecting} and Video Highlight Detection~\cite{liu2022umt}, their practicality is limited as they cannot handle multiple VTG subtasks concurrently and require extensive task-specific fine-tuning to achieve adequate performance.
More recently, Video Large Language Models (Video-LLMs) \cite{xu2024pllava,li2024mvbench,cheng2024videollama,lin2023video,wang2024qwen2} have made remarkable advancements in \textit{coarse-grained} video content understanding (e.g., VideoQA \cite{li2024mvbench,fu2024video}), yet they struggle to extend such capabilities to tasks requiring \textit{fine-grained} temporal localization (e.g., VTG \cite{zhou2018towards,gao2017tall}), due to the lack of dedicated mechanisms for explicitly modeling temporal boundaries.

Compared to generalist Video-LLMs, VTG-specific models that encode temporal tokens~\cite{huang2024lita,guo2024vtg,wang2024grounded} or are trained on temporally-annotated instruction-tuning datasets~\cite{ren2024timechat,wang2024hawkeye,huang2024vtimellm} demonstrate marginal improvements, as they blend temporal information with text tokens \textit{without explicitly modeling} timestamps and saliency scores, despite videos inherently possessing structured information beyond textual descriptions~\cite{guo2024trace}.
Furthermore, existing VTG-specific Video-LLMs rely exclusively on a single LLM as their reasoning backbone, \textit{processing all task tokens indiscriminately} with shared parameters. This parameter-sharing paradigm inherently constrains the model's ability (\Cref{fig:teaser}-Right) to specialize in the diverse subtasks of VTG—such as timestamp prediction, saliency estimation, and caption generation—each of which may demand distinct feature representations for optimal performance.

\noindent{\bf Proposed Method.}
To address the misalignment between language modeling of existing Video-LLMs and inherent video temporal structure, we introduce \textsl{TimeExpert}, an event-centric modeling framework that decouples the prediction of timestamps, saliency scores, and textual descriptions into distinct specialized tasks, organized in an interleaved sequence.
To precisely model the varying token importance across these decoupled tasks, we replace the single LLM backbone of existing Video-LLMs with a fine-grained Mixture-of-Experts (MoE) decoder, with our proposed token-aware dynamic gating and token-adaptive routing strategies, which achieves both computational efficiency and modeling precision by activating only task-relevant parameters during both training and inference.
To harness each expert's potential, we develop a task-dependent auxiliary loss that encourages more frequently activated experts to process a greater proportion of task-relevant tokens.

\noindent{\bf Our Contributions.}
\textbf{(1):} To our knowledge, TimeExpert is the first to explore the task token importance in VTG by introducing a MoE-based framework with dynamic expert routing.
Our expert-guided architecture allows for adaptive learning across different VTG subtasks, effectively mitigating task interference while leveraging expert specialization.
\textbf{(2):} We conduct extensive experiments across multiple VTG tasks on several benchmarks to demonstrate the effectiveness of TimeExpert. 
In particular, we demonstrate substantial improvements over state-of-the-art Video-LLMs through quantitative results and ablation studies, highlighting the advantages of expert-driven token specialization.

\begin{figure}[t]
    \centering
    \includegraphics[width=\linewidth]{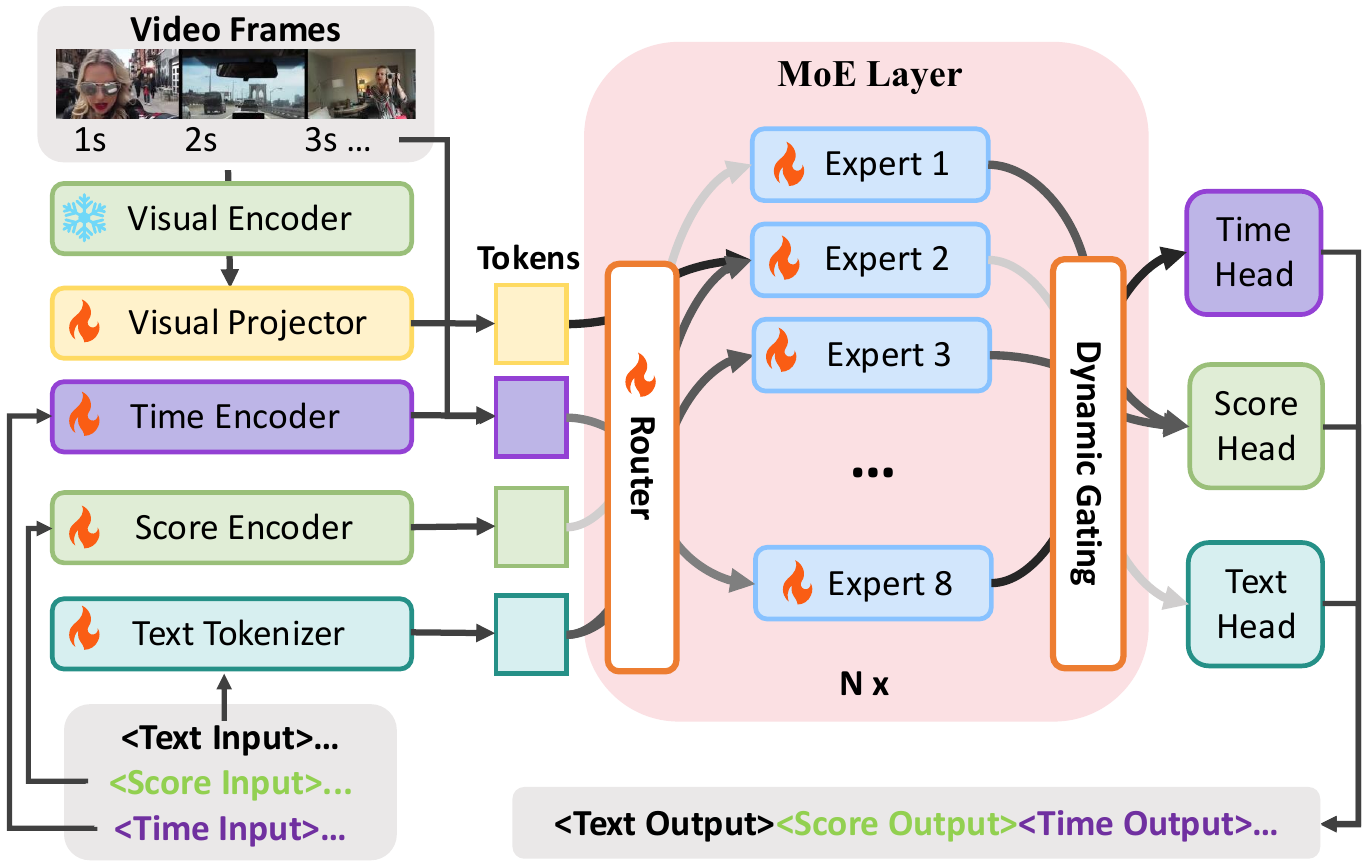}
    \caption{
    \textbf{Architecture Overview of \textsl{TimeExpert}.} Our model leverages independent encoders and decoding heads to process time, score, and text inputs and outputs. The timestamps and saliency scores of sampled frames are encoded into special tokens and integrated into the corresponding visual tokens. During inference, the generated response follows a structured format, sequentially incorporating time tokens, score tokens, and text tokens.}
    \label{fig:framework}
\end{figure}

%% file: sec/2_related.tex
\section{Related Work}
\label{sec:related}

{\bf Video Temporal Grounding}.
Video Temporal Grounding (VTG) aims to precisely localize relevant temporal boundaries in a video given textual queries \cite{lin2023univtg}.
It can be divided into subtasks like Moment Retrieval \cite{gao2017tall,oncescu2021queryd}, Dense Video Captioning \cite{caba2015activitynet, zhou2018towards,tang2019coin}, Video Summarization \cite{song2015tvsum,gygli2014creating}, and Video Highlight Detection \cite{lei2021detecting,liu2022umt}.
Traditional methods primarily rely on large-scale video-text pretraining, using training objectives such as video-text contrastive learning \cite{xu2021videoclip,wang2022internvideo}, video-text matching \cite{chen2023vast,li2023unmasked}, and timestamp regression \cite{moon2023correlation,moon2023query,zeng2024unimd,zala2023hierarchical}.
Despite their initial success, these methods are constrained by their generalization ability to handle only a single task per model and require substantial fine-tuning for each new downstream application.

\noindent{\bf Video-LLMs for VTG Tasks.}
With the growing reasoning capabilities of Video-LLMs \cite{xu2024pllava,li2024mvbench,cheng2024videollama,lin2023video,wang2024qwen2}, numerous studies have investigated their adaptation to VTG tasks (\Cref{fig:method}-(a)).
For example, TimeChat \cite{ren2024timechat}, VTimeLLM \cite{huang2024vtimellm}, and HawkEye \cite{huang2024lita} fine-tune the Video-LLMs on refined temporal-annotated instruction datasets.
Following that, several studies further enhance the Video-LLM's understanding of timestamps by introducing special tokens \cite{huang2024lita, guo2024vtg} or a temporal perception module \cite{qian2024momentor}. 
However, these methods adopt a purely text-to-text framework, formulating VTG as a natural language generation problem, which lacks explicit temporal modeling.
Recently, TRACE \cite{guo2024trace} proposes a causal event modeling framework to capture the inherent structure of videos.
Nevertheless, it still fails to incorporate task-specific properties, as they process all task tokens within a single shared reasoning backbone, leading to task interference and limited specialization.
In contrast, TimeExpert simultaneously structures Video-LLM outputs into event representations and leverages adaptive expert specialization, enhancing both efficiency and accuracy across diverse VTG tasks.

\noindent{\bf LLMs with Mixture-of-Experts.}
The Mixture-of-Experts (MoE)~\cite{eigen2013learning, shazeer2017outrageously, li2025uni} architecture (\Cref{fig:method}-(b)) has been widely adopted to improve the efficiency of large-scale foundation models during both training and inference by replacing feed-forward network layers in transformer models with multiple specialized sub-networks, known as experts, which enables sparse parameter activation.
The activated experts are selected by a gating network, which contain a pre-defined router.
Prior methods often implement static top-$k$ routing across all tokens, several studies~\cite{yang2024xmoe, zeng2024adamoe, huang2024harder,guo2024dynamic} explored on dynamic routing mechanism that allow tokens to activate varying numbers of experts.
However, these approaches either introduce additional hyperparameter complexity or demonstrate limited validation confined to text-to-text tasks, constraining their flexibility and cross-task generalizability.
In contrast, our proposed \textsl{TimeExpert} (\Cref{fig:method}-(c)) introduces dynamic expert routing tailored for Video-LLMs through adaptively adjusting the number of activated experts based on task token importance, ensuring effective expert specialization while eliminating reliance on rigid hyperparameter tuning.

%% file: sec/3_method.tex
\section{TimeExpert: An Expert-guided Video LLM}
\label{sec:method}

In this section, we firstly introduce some fundamental preliminaries in \Cref{sec-31}. 
Then, we propose \textsl{TimeExpert} (\Cref{fig:framework}), which integrates a dynamic expert selection mechanism with a structured event modeling framework. Our approach implements task-aware dynamic gating (\Cref{sec-32}) and token-adaptive routing (\Cref{sec-33}) to selectively activates specialized experts based on token relevance across diverse tasks.
In \Cref{sec-34}, we discuss our proposed task-dependent auxiliary loss that optimizes expert utilization and thereby enhances the stability of MoE-based learning.
In \Cref{sec-35}, we elaborate on our training recipe.

\subsection{Preliminaries}
\label{sec-31}
\noindent\textbf{Problem Definition.} 
Given a textual instruction $\mathcal{I}$ and a sequence of video frames $\mathcal{F}$ from a video, the objective of VTG task is to generate a structured event representation $\mathcal{R}$. 
The output sequence $\mathcal{R}$ consists of discrete events $\{\epsilon_1, \epsilon_2, \dots, \epsilon_M\}$, where each event $\epsilon_m$ encapsulates three key components: a timestamp $t_m$, a saliency score $s_m$, and a textual caption $c_m$. Formally, we have:
{\small
\begin{equation}
    \mathcal{R} = \{\epsilon_1, \epsilon_2, \dots, \epsilon_M\} = \{(t_m, s_m, c_m) \mid 1 \leq m \leq M\}.
    \label{eq1}
\end{equation}
}

\noindent\textbf{Causal Event Modeling.} 
Following \cite{guo2024trace}, we formulate the event prediction process in a causal manner, leveraging the sequential dependencies between events.
Specifically, given the past event sequence $\epsilon_{1:m-1}$, the textual instruction $\mathcal{I}$, and the video frames $\mathcal{F}$, the probability distribution of the next event $\epsilon_m$ is defined as:
{\small
\begin{equation}
    \mathcal{P}(\epsilon_m \mid \epsilon_{1:m-1}, \mathcal{I}, \mathcal{F}) = \mathcal{P}(t_m, s_m, c_m \mid \epsilon_{1:m-1}, \mathcal{I}, \mathcal{F}).
    \label{eq2}
\end{equation}
}
This causal formulation ensures that event generation follows a structured temporal order, where each event is progressively determined based on prior events, textual cues, and visual content.

\subsection{Task-aware Dynamic Gating}
\label{sec-32}
\textbf{Vanilla MoEs.} (Figure~\ref{fig:method}-(b)) employ the top-$k$ gating strategy, which activates a fixed number of experts for each input token, determined by computing gating scores via a learned network $g$.
Specifically, for a given token embedding $\mathbf{x} \in \mathbb{R}^{d}$, the gating function follows:
\begin{equation}
    g(\mathbf{x}) \in \mathbb{R}^{K} := \text{softmax}(\mathbf{W}_g^T \mathbf{x}) ,
\end{equation}
where $\mathbf{W}_g \in \mathbb{R}^{d \times K}$ represents the gating network parameters, and $K$ denotes the total number of experts. Then, the output of this MoE layer is then formulated as:
\begin{equation}
    \mathbf{y} = \frac{1}{\sum_{e \in \text{Top-}k(g(\mathbf{x}))} g(\mathbf{x})_e} \sum_{e \in \text{Top-}k(g(\mathbf{x}))} g(\mathbf{x})_e \mathbf{E}_e(\mathbf{x}),
\end{equation}
where $\mathbf{E}_e(\mathbf{x}) \in \mathbb{R}^{d}$ is the output of the $e$-th expert given input $\mathbf{x}$, and $g(\mathbf{x})_e$ is the corresponding gating score.

\begin{figure}[ht]
    \centering
    \includegraphics[width=\linewidth]{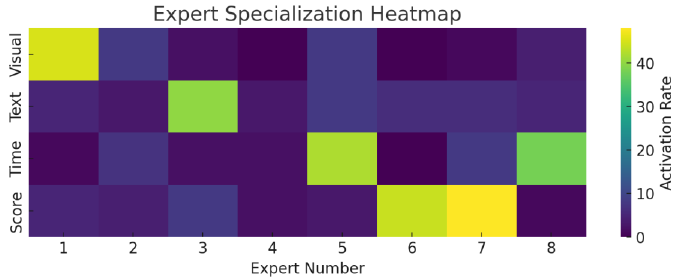}
    \caption{
    \textbf{Visualization of Expert Assignments on Various Task Tokens using Our Vanilla MoE Implementation.}
    We take layer 4 as an example and  only visualized the first 8 experts out of a total of 64, due to the space limitation.
    }
    \label{fig:act_rate}
\end{figure}

\noindent\textbf{Implicit Task Preference.} 
Vanilla MoEs face some limitations, such as
\textbf{(1):} the reliance on a predefined $k$ for expert selection leads to suboptimal flexibility \cite{yang2021m6,clark2022unified,fan2024towards}, (as task complexity varies, different tokens may require a varying number of experts based on their importance); and
\textbf{(2):} the standard gating function treats all tokens uniformly, neglecting the inherent differences in task-specific significance, which may lead to inefficient expert routing and task interference \cite{huang2024harder}.
Surprisingly, we observe an \textit{implicit task preference} in these models on VTG tasks, as visualized in \Cref{fig:act_rate} of expert assignments.
In particular, we measure the ratio of activated task-specific tokens to the number of processed text tokens in that layer.
Our findings reveal that certain experts, such as expert \#7, consistently activate in response to specific subtasks (e.g., score tokens), \textit{even though the model was not explicitly trained to establish such specialization}, implying an inherent task-driven preference.

\noindent\textbf{Explicit Task-Aware Gating.} 
Based on this observation, we propose a task-aware dynamic gating mechanism that jointly considers token-level probabilities and task-specific activation statistics.
Instead of applying a static threshold, we introduce a \textit{task-weighted gating function} that adjusts expert activation dynamically by incorporating task token activation rates. Formally, given an expert representation matrix $\mathbf{W}_g \in \mathbb{R}^{d \times K}$ and an input token $\mathbf{x} \in \mathbb{R}^{d}$, we define:

\begin{equation}
    s(\mathbf{x}) = \text{cos}(\mathbf{x}, \mathbf{W}_g), 
\end{equation}
\begin{equation}
    g(\mathbf{x}) = \text{sign}\left(\sigma\left(\frac{s(\mathbf{x}) + \alpha A_t}{1 + \alpha} \right) - \sigma(\mathbf{G}) \right),
\end{equation}
where $A_t$ represents the historical activation rate of the task token type associated with $\mathbf{x}$; $\alpha$ is a scaling coefficient that controls the influence of task importance; $\sigma(\cdot)$ denotes the sigmoid function for score normalization; $\mathbf{G} \in \mathbb{R}^K$ is a learnable threshold where a token routes to expert $e$ only if its similarity score $s(\mathbf{x}) \in \mathbb{R}^K$ exceeds $\mathbf{G}$.
Following \cite{guo2024dynamic}, we make the binary \emph{sign()} differentiable by copying the upstream gradient of $g(\mathbf{x})$ to the pre-activation term $\sigma(s(\mathbf{x}, A_t))-\sigma(\mathbf{G})$.
With $A_t$, the model \textit{explicitly prioritizes} experts that have demonstrated effectiveness for specific task tokens, encouraging specialization while ensuring under-utilized experts do not get excessively activated.

\begin{table*}[ht]
    \centering
    \renewcommand{\arraystretch}{1.2}
    \small
    \resizebox{0.8\textwidth}{!}{
        \begin{tabular}{l|c|c}
            \toprule
            \textbf{Training Stage} & \textbf{Datasets} & \textbf{No. of Samples} \\
            \midrule
            \multirow{1}{*}{Stage 1: Task Module Pretraining} &  
            \makecell{Valley \cite{luo2023valley}, LLaVA-Image \cite{liu2023visual}, TextVR \cite{wu2025large}, \\ ShareGPT4Video \cite{chen2024sharegpt4video}, VTG-IT \cite{guo2024vtg}} & 
            \multirow{1}{*}{1.9M} \\
            \midrule
            \multirow{1}{*}{Stage 2: MoE Decoder Pretraining} &  
            \makecell{$\text{Valley}^\ast$, $\text{TextVR}^\ast$, $\text{ShareGPT4Video}^\ast$, $\text{VTG-IT}^\ast$, \\ ActivityNet Captions \cite{caba2015activitynet}, VideoChatGPT \cite{maaz2023video}, \\ InternVid \cite{wang2022internvideo}, Next-QA \cite{xiao2021next}} & 
            \multirow{1}{*}{0.9M} \\
            \midrule
            \multirow{1}{*}{Stage 3: Supervised Fine-tuning} &  
            \makecell{Filtered and Re-annotated Data from: \\ Previous Stages' Data, EgoQA \cite{nguyen2024encoding}, STAR \cite{wu2024star}, \\ Moment-10M \cite{qian2024momentor}, LLaVA-Video-178K \cite{zhang2024video}} & 
            \multirow{1}{*}{2.3M} \\
            \bottomrule
        \end{tabular}
    }
    \caption{\textbf{Training Data Recipe of \textsl{TimeExpert}.} These data are categorized into three training stages. $\ast$ denotes we utilize the filtered subset of the original dataset.}
    \label{tab:vtg_moe_data_recipe}
\end{table*}

\subsection{Token-adaptive Routing}
\label{sec-33}
In this section, we introduce a token-adaptive routing mechanism (\Cref{fig:method}-(c)) that dynamically facilitates expert addition and removal based on recorded activation trends. The routing mechanism comprises three key components, including
\textbf{(1):} Task-specific Routing Recording: We continuously monitor expert utilization across different task tokens and the routing results during training.
\textbf{(2):} Adaptive Expert Addition: New experts are allocated when certain task tokens persistently fail to activate appropriate experts, and
\textbf{(3):} Redundant Expert Removal: Experts with persistently low activation rates are pruned to maintain efficiency.

\noindent\textbf{Task-specific Routing Recording.} Instead of solely tracking which experts get activated, we maintain a per-task token activation frequency metric.
Specifically,
(1) For each expert $e$, we log its activation timestamps, denoted as $\mathbf{R}_E \in \mathbb{R}^{K}$;
(2) For input tokens that fail to activate any expert, we compute their aggregated embeddings as $\mathbf{R}_S \in \mathbb{R}^{d}$;
(3) We maintain an activation rate tracker $A_t$ per task token type, which influences expert selection in subsequent iterations.

\noindent\textbf{Adaptive Expert Addition.} When a significant proportion of task tokens fail to activate any expert, we introduce new experts whose representations are initialized as:
\begin{equation}
    \mathbf{W}_{g, K+1} = \frac{\mathbf{R}_S}{||\mathbf{R}_S||}, \quad \mathbf{G}_{K+1} = 0,
\end{equation}
where $\mathbf{R}_S$ captures the average embedding of underrepresented task tokens. This ensures that newly added experts are aligned with the missing task representations.

\noindent\textbf{Redundant Expert Removal.} Experts are pruned if their activation rate $A_e$ remains below a threshold $\tau_{\text{min}}$:
\begin{equation}
    \mathcal{E}_{\text{remove}} = \{ e \mid A_e < \tau_{\text{min}} \}.
\end{equation}
This prevents redundant experts from occupying parameter space while ensuring that experts specializing in low-frequency but important task tokens are retained.

\subsection{Task-dependent Auxiliary Loss}
\label{sec-34}
To enhance the efficiency and specialization of expert activation in TimeExpert, we introduce a task-dependent auxiliary loss, which ensures that experts with higher activation rates receive more relevant task tokens while preventing excessive expert overlap.

Existing auxiliary loss functions \cite{fedus2022switch,wu2024multi} primarily focus on load balancing among experts, which may conflict with our goal of leveraging expert specialization for VTG tasks.
Instead, our auxiliary loss promotes a dynamic allocation strategy by reinforcing the association between frequently activated experts and their corresponding task tokens while preventing under-utilization or excessive redundancy.

Our auxiliary loss consists of two key components: \emph{task-aware concentration term} (Left) and \emph{activation regularization term} (Right).
The task-aware concentration loss ensures that experts specialize in processing specific types of task tokens (e.g., timestamps, saliency scores, captions) by amplifying the assignment of relevant tokens to frequently activated experts.
The activation regularization loss prevents the over-activation of a single expert.
The two loss terms work synergistically to ensure a diverse and efficient allocation of task tokens. In summary, we define our auxiliary loss as follows:
\begin{equation}
\resizebox{0.39\textwidth}{!}{$
\mathcal{L}_{\mathrm{aux}} = \lambda_1 \sum_{e=1}^K \left( \frac{A_e}{\sum_{j=1}^K A_j} - \frac{N_e}{\sum_{j=1}^K N_j} \right)^2
+ \lambda_2 \sum_{e=1}^K \|\mathbf{w}_{g,e}\|_2^2
$}
\label{eq:aux}
\end{equation}
where $A_e$ represents the activation count of expert $e$ over a batch, $N_e$ denotes the number of task tokens assigned to expert $e$, $\mathbf{w}_{g,e}$ is the representation vector of expert $e$, and $\lambda_1$ and $\lambda_2$ are hyperparameters controlling the balance between task-aware specialization and numerical stability.

\subsection{Training Recipe}
\label{sec-35}

\noindent\textbf{Training Data.} To ensure comprehensive learning across various video-language understanding tasks, we curate a diverse dataset spanning three progressive training stages, as summarized in \Cref{tab:vtg_moe_data_recipe}.
These utilized public datasets cover a wide range of temporal reasoning, video grounding, and captioning challenges.
We conducted systematic processing, filtering, and re-annotating techniques that remove inaccurate temporal annotations, ensuring high-quality labels.

\noindent\textbf{Training Strategy.}
The training process of TimeExpert includes three stages:
(1) \emph{Stage 1 - Task Module Pretraining.} In the first stage, we pretrain the task-specific modules, including the vision compression layer, task encoder, and task heads, using 1.9M general multimodal video-text samples. This stage ensures that the model learns fundamental video representation capabilities before incorporating MoE decoder.
(2) \emph{Stage 2 - MoE Decoder Pretraining.} Before fine-tuning the full model, we introduce an intermediate MoE pretraining stage with 0.9M samples to ensure that expert routing aligns with different VTG-specific task tokens. This step mitigates the risk of expert collapse (i.e., all tokens being routed to the same expert) and enhances task-aware expert specialization.
(3) \emph{Stage 3 - Supervised Fine-tuning.} In the final stage, we fine-tune the entire TimeExpert framework on 2.3M samples, allowing both the task modules and the MoE decoder to be optimized jointly.

\noindent\textbf{Training Objectives.}
During the first stage, we adopt cross-entropy loss for each generated token to teach basic video-event formatting.
During MoE Decoder Pretraining, besides the cross-entropy loss, we apply $z$-loss \cite{zoph2022st} to stabilize training and proposed task-aware auxiliary loss to mitigate expert collapse and drive specialization.
During Supervised Fine-tuning, we fine-tune the full model (except visual encoder) using same trio of objectives, ensuring stable routing while the model learns task-specific instructions.

%% file: sec/4_exp.tex
\section{Experiments}
\label{sec:exp}

\begin{table*}[t]
    \centering
    \small
    \renewcommand{\arraystretch}{1.2}
    \begin{adjustbox}{width=\linewidth}
        \begin{tabular}{l|c|ccc|cc|cc}
            \toprule
            \multirow{3}{*}{\centering\textbf{Method}} &
            \multirow{3}{*}{\makecell[c]{\textbf{No.\ of}\\\textbf{Activated}\\\textbf{Parameters}}} &
            \multicolumn{3}{c|}{\textbf{Dense Video Captioning}} &
            \multicolumn{2}{c|}{\textbf{Moment Retrieval}} &
            \multicolumn{2}{c}{\textbf{Video Highlight Detection}} \\[0.25em]
            & & \multicolumn{3}{c|}{(YouCook2 \cite{zhou2018towards})} &
            \multicolumn{2}{c|}{(Charades-STA \cite{gao2017tall})} &
            \multicolumn{2}{c}{(QVHighlights \cite{lei2021detecting})} \\ \cline{3-9}
            & & SODA$_c$ ($\uparrow$) & CIDEr ($\uparrow$) & F1 Score ($\uparrow$) &
            R@1$_{\text{IoU}=0.5}$ ($\uparrow$) & R@1$_{\text{IoU}=0.7}$ ($\uparrow$) &
            mAP ($\uparrow$) & HIT@1 ($\uparrow$) \\
            \midrule
            TimeChat \cite{ren2024timechat}           & 7B & 1.2 & 3.4 & 12.6 & 32.2 & 13.4 & 14.5 & 23.9 \\
            VTimeLLM \cite{huang2024vtimellm}         & 7B & --  & --  & --   & 27.5 & 11.4 & --   & --   \\
            Momentor \cite{qian2024momentor}          & 7B & --  & --  & --   & 26.6 & 11.6 &  7.6 & --   \\
            HawkEye \cite{wang2024hawkeye}            & 7B & --  & --  & --   & 31.4 & 14.5 & --   & --   \\
            VTG-LLM \cite{guo2024vtg}                 & 7B & 1.5 & 5.0 & 17.5 & 33.8 & 15.7 & 16.5 & 33.5 \\
            TRACE \cite{guo2024trace}                 & 7B & 2.2 & \underline{8.1} & 22.4 & 40.3 & 19.4 & 26.8 & 42.7 \\
            \rowcolor{gray!20}
            TimeExpert (adaptive $k$)                 & $\approx$ 5.9B / 3.5B / 4.8B & \textbf{2.5} & \textbf{8.2} & \textbf{23.6} & \textbf{42.8} & \textbf{20.3} & \textbf{29.6} & \textbf{46.9} \\
            \rowcolor{gray!20}
            TimeExpert (TRACE's data)                 & $\approx$ 5.2B / 3.1B / 4.0B & \underline{2.4} & \underline{8.1} & \underline{23.3} & \underline{41.9} & \underline{20.1} & \underline{29.1} & \underline{46.3} \\
            \bottomrule
        \end{tabular}
    \end{adjustbox}
    \caption{\textbf{Zero-shot Performance Comparison of \textsl{TimeExpert} against several state-of-the-art VTG-specific Video-LLMs on Dense Video Captioning, Temporal Grounding, and Video Highlight Detection.} Some results are sourced from \cite{guo2024trace}. The best results are in \textbf{bold}. We also \underline{underline} the second-best results.}
    \label{tab:video_captioning}
\end{table*}

\subsection{Experimental Details}

\begin{table*}[t]
    \centering
    \small
    \renewcommand{\arraystretch}{1.2}
    \begin{tabular}{l|ccc|cc}
    \toprule
    \multirow{3}{*}{\textbf{Method}} &
    \multicolumn{3}{c|}{\textbf{Dense Video Captioning}} &
    \multicolumn{2}{c}{\textbf{Moment Retrieval}} \\[0.15em]
    & \multicolumn{3}{c|}{(YouCook2 \cite{zhou2018towards})} &
    \multicolumn{2}{c}{(Charades-STA \cite{gao2017tall})} \\ \cline{2-6}
    & SODA$_c$ ($\uparrow$) & CIDEr ($\uparrow$) & F1 Score ($\uparrow$)
    & R@1$_{\text{IoU}=0.5}$ ($\uparrow$) & R@1$_{\text{IoU}=0.7}$ ($\uparrow$) \\
    \midrule
    \multicolumn{6}{l}{\textbf{\textit{Traditional Models}}} \\
    \midrule
    PDVC \cite{wang2021end}                   & 4.4 & 22.7 & --  & --   & --   \\
    Vid2Seq \cite{yang2023vid2seq}            & 5.7 & 25.3 & 23.5 & --   & --   \\
    Vid2Seq (Audio Input) \cite{yang2023vid2seq} & \textit{7.9} & \textit{47.1} & \textit{27.3} & -- & -- \\
    CM$^2$ \cite{kim2024you}                  & 5.3 & 31.7 & 28.4 & --   & --   \\
    \midrule
    \multicolumn{6}{l}{\textbf{\textit{VTG-specific Video-LLMs}}} \\
    \midrule
    TimeChat \cite{ren2024timechat}           & 3.4 & 11.0 & 19.5 & 46.7 & 23.7 \\
    VTG-LLM \cite{guo2024vtg}                 & 3.6 & 13.4 & 20.6 & 57.2 & 33.4 \\
    TRACE \cite{guo2024trace}                & 6.7 & 35.5 & 31.8 & 61.7 & 41.4 \\
    \rowcolor{gray!20}
    TimeExpert (Ours)                         & \textbf{7.2} & \textbf{39.0} & \textbf{33.5} & \textbf{64.1} & \textbf{43.3} \\
    \bottomrule
    \end{tabular}
    \caption{\textbf{Fine-tuned Performance of \textsl{TimeExpert}.} Some results are from \cite{guo2024trace}. We report the performance of each model after fine-tuning on YouCook2 and Charades-STA datasets. We \textbf{bold} the best scores and \textit{italicize} results that benefit from additional audio inputs.}
    \label{tab:finetune}
\end{table*}

\begin{table*}[t]
    \centering
    \small
    \renewcommand{\arraystretch}{1.2}
    \begin{tabular}{lcccc|ccc}
        \toprule
        \multirow{2}{*}{\textbf{Method}} &
        \multicolumn{4}{c|}{\textbf{Dense Video Captioning}} &
        \multicolumn{3}{c}{\textbf{Moment Retrieval}} \\
        \cmidrule(lr){2-5} \cmidrule(lr){6-8}
        & METEOR ($\uparrow$) & SODA$_c$ ($\uparrow$) & CIDEr ($\uparrow$) & F1 Score ($\uparrow$) &
          R@1$_{\text{IoU}=0.5}$ ($\uparrow$) & R@1$_{\text{IoU}=0.7}$ ($\uparrow$) & mIoU ($\uparrow$) \\
        \midrule
        TimeChat \cite{ren2024timechat}          & 5.7 & 4.7 & 19.0 & 36.9 &  4.6 &  2.0 &  6.9 \\
        VTimeLLM \cite{huang2024vtimellm}        & 6.8 & 5.8 & 27.6 & --   & 29.5 & 14.2 & 31.4 \\
        Momentor$^{*}$ \cite{qian2024momentor}   & 4.7 & 2.3 & 14.9 & --   & 23.0 & 12.4 & 29.3 \\
        VTG-LLM \cite{guo2024vtg}               & 5.9 & 5.1 & 20.7 & 34.8 &  8.3 &  3.7 & 12.0 \\
        TRACE \cite{guo2024trace}               & 6.4 & 6.0 & 25.9 & 39.3 & 37.7 & 24.0 & 39.0 \\
        \rowcolor{gray!20}
        TimeExpert (Ours)                       & \textbf{7.0} & \textbf{6.5} & \textbf{28.4} & \textbf{40.5} &
                                                   \textbf{39.2} & \textbf{26.1} & \textbf{41.5} \\
        \bottomrule
    \end{tabular}
    \caption{\textbf{Performance Comparison on ActivityNet Captions \cite{krishna2017dense}.} Some results are sourced from \cite{guo2024trace}. * indicates zero-shot evaluation. The best results are in \textbf{bold}. TimeExpert achieves state-of-the-art results across both \textbf{Dense Video Captioning} and \textbf{Moment Retrieval}.}
    \label{tab:anet_results}
\end{table*}

\begin{table*}[t]
\centering
\small
\renewcommand{\arraystretch}{1.2}
\begin{adjustbox}{width=\linewidth}
\begin{tabular}{l|ccc|cc|cc}
\toprule
\multirow{3}{*}{\textbf{Method}} 
        & \multicolumn{3}{c|}{\textbf{Dense Video Captioning}} 
        & \multicolumn{2}{c|}{\textbf{Moment Retrieval}} 
        & \multicolumn{2}{c}{\textbf{Video Highlight Detection}} \\[0.15em]
        & \multicolumn{3}{c|}{(YouCook2 \cite{zhou2018towards})} 
        & \multicolumn{2}{c|}{(Charades-STA \cite{gao2017tall})} 
        & \multicolumn{2}{c}{(QVHighlights \cite{lei2021detecting})} \\ \cline{2-8}
        & SODA$_c$ ($\uparrow$) & CIDEr ($\uparrow$) & F1 Score ($\uparrow$) 
        & R@1$_{\text{IoU}=0.5}$ ($\uparrow$) & R@1$_{\text{IoU}=0.7}$ ($\uparrow$) 
        & mAP ($\uparrow$) & HIT@1 ($\uparrow$) \\
\midrule
\multicolumn{8}{c}{\textbf{\textit{Ablation Studies on Design Choices}}} \\
\midrule
w/o token-adaptive routing      & 2.1 & 7.6 & 22.2 & 40.5 & 19.2 & 26.5 & 42.6 \\
$\text{w/o separate encoders}^\ast$ & -   & -   & -    & -    & -    & -    & -    \\
w/o task-dependent loss         & 2.4 & 7.9 & 22.8 & 41.3 & 19.7 & 28.1 & 45.2 \\
\rowcolor{gray!20}
TimeExpert (Ours)               & \textbf{2.5} & \textbf{8.2} & \textbf{23.6} & \textbf{42.8} & \textbf{20.3} & \textbf{29.6} & \textbf{46.9} \\
\midrule
\multicolumn{8}{c}{\textbf{\textit{Ablation Studies on No. of Input Frames}}} \\
\midrule
TimeExpert (8 frames)           & 1.5 & 5.1 & 18.9 & 30.2 & 13.9 & 22.1 & 38.4 \\
TimeExpert (64 frames)          & 2.1 & 7.5 & 21.9 & 37.5 & 17.8 & 27.2 & 44.9 \\
TimeExpert (128 frames)         & \textbf{2.5} & \textbf{8.2} & \textbf{22.0} & \textbf{41.5} & \textbf{19.9} & \textbf{29.3} & \textbf{46.5} \\
\midrule
\multicolumn{8}{c}{\textbf{\textit{Ablation Studies on No. of Activated Experts}}} \\
\midrule
Vanilla MoE ($k{=}2$)           & 2.3 & 8.1 & 22.8 & 42.1 & 19.8 & 29.2 & 45.8 \\
Vanilla MoE ($k{=}4$)           & 2.4 & \textbf{8.2} & 23.3 & 42.5 & 20.1 & 29.5 & 46.6 \\
Vanilla MoE ($k{=}6$)           & \textbf{2.5} & \textbf{8.2} & \textbf{23.5} & \textbf{42.8} & \textbf{20.2} & 29.6 & \textbf{46.9} \\
Vanilla MoE ($k{=}8$)           & \textbf{2.5} & 8.1 & 23.4 & \textbf{42.8} & \textbf{20.2} & \textbf{29.8} & 46.8 \\
\bottomrule
\end{tabular}
\end{adjustbox}
\caption{\textbf{Ablation Study Results of \textsl{TimeExpert}.} We evaluate the impact of removing key components, varying the number of input frames, and adjusting the number of activated experts. Results under the default setting are highlighted in \colorbox{gray!20}{gray}.  \(\ast\) denotes that directly processing time tokens and score tokens with the text tokenizer causes the model to fail to follow instructions.}
\label{tab:ablation_vtg_moe}
\end{table*}

{\bf Implementation Details.}
We adopt ARIA \cite{li2024aria} as the base MoE Video-LLM.
Specifically, we utilize a lightweight visual encoder with 438M parameters, consisting of a Vision Transformer (ViT) \cite{dosovitskiy2020image} and a projection module. The ViT weights are initialized from \cite{zhai2023sigmoid}.
Each video frame is initially encoded into 128 (for medium resolution) or 256 (for high resolution) visual tokens, which are then compressed via slot‑based token compression \cite{guo2024vtg} down to 8 visual tokens per frame.
The time encoder and score encoder share the same architecture, both initialized with a tokenizer containing 11 number tokens for representing timestamps/scores, a separator token marking the end of each timestamp/score, and a switching token indicating task transitions.
Specifically, following \cite{guo2024trace}, we insert the token $\langle sep \rangle$ between consecutive timestamps or scores and append $\langle sync \rangle$ at the end of the sequence.

\noindent{\bf Comparison Methods.}
We select TimeChat \cite{ren2024timechat}, VTimeLLM \cite{huang2024vtimellm}, Momentor \cite{qian2024momentor}, HawkEye \cite{wang2024hawkeye}, VTG-LLM \cite{guo2024vtg}, and TRACE \cite{guo2024trace} as comparison methods.

\noindent{\bf Evaluation Benchmarks.}
To thoroughly evaluate TimeExpert in Video Temporal Grounding, we assess it across three VTG tasks: Dense Video Captioning, Moment Retrieval, and Video Highlight Detection, utilizing datasets such as YouCook2 \cite{zhou2018towards}, Charades-STA \cite{gao2017tall}, QVHighlights \cite{lei2021detecting}, and ActivityNet Captions \cite{krishna2017dense}.

\noindent{\bf Evaluation Metrics.}
For Dense Video Captioning, we use metrics including $\text{SODA}_\text{c}$ \cite{fujita2020soda}, which is specifically tailored for the video's storyline; CIDEr score \cite{vedantam2015cider}, which measures consensus between generated and reference captions by computing the TF-IDF weighted n-gram similarity across multiple reference captions; F1 Score, which quantifies the model's capacity on precisely locating timestamps; and METEOR \cite{banerjee2005meteor}, which combines uni-gram precision and recall into a harmonic mean while incorporating stemming, synonym matching, and a fragmentation penalty so that longer contiguous matches are rewarded.
For Moment Retrieval, we compute the Intersection over Union (IoU) between the timestamps predicted by the model and ground truth, including Recall at IoU thresholds of \{0.5, 0.7\} and mean IoU.
For Video Highlight Detection, we calculate both the mean Average Precision (mAP) at IoU thresholds of \{0.5, 0.75\}, and HIT@1, which represents the hit ratio of the highest scored clip.

\subsection{Quantitative Results}

\noindent\textbf{Zero-shot Performance on VTG Tasks.}
In \Cref{tab:video_captioning}, we present the zero-shot performance of TimeExpert against several state-of-the-art baselines on three widely adopted VTG benchmarks.
We report the number of activated parameters for a transparent comparison.
Across the three benchmarks, the adaptive-$k$ variant activated an average of 14.5, 9.1, and 11.8 experts, whereas the TRACE-trained variant averaged 13.5, 8.3, and 10.5.
Notably, our approach outperforms other VTG-specific Video-LLMs across all three benchmarks with fewer activated parameters.
It achieves a 0.3\% and 1.2\% performance gain on YouCook2 \cite{zhou2018towards} using the $\text{SODA}_\text{c}$ and F1 Score metrics, respectively;
a 2.5\% and 0.9\% performance boost in Recall with IoU = \{0.5, 0.7\} thresholds on Charades-STA \cite{gao2017tall};
and a 2.8\% and 4.2\% performance gain for the mAP and HIT@1 metrics on QVHighlights \cite{lei2021detecting}.
This validates that dynamic gating and adaptive routing not only ensures efficient expert utilization but also fully leverages task specialization.

\noindent\textbf{Fine-tuned Performance on VTG Tasks.} In \Cref{tab:finetune}, we present the results of fine-tuning TimeExpert on YouCook2 and Charades-STA datasets for 2 epochs. The results indicate that:
(1) \textit{TimeExpert surpasses both generalist and task-specific baselines.} Unlike TimeChat and VTG-LLM, which struggle to fully benefit from fine-tuning, our MoE-based approach effectively adapts to downstream datasets, achieving new state-of-the-art results.
Notably, it surpasses TRACE in both Moment Retrieval and Dense Video Captioning tasks, highlighting the benefits of dynamic expert activation in video-language modeling.
(2) \textit{TimeExpert enhances retrieval and captioning tasks simultaneously.} Our model achieves a +3.5 improvement in CIDEr and a +1.7 boost in F1 Score on YouCook2, alongside an absolute gain of 2.4 in $\text{R@}1_{\text{IoU}=0.5}$ and 1.9 in $\text{R@}1_{\text{IoU}=0.7}$ on Charades-STA.
These results demonstrate that TimeExpert's token-adaptive routing effectively captures event structures, enabling improved video reasoning.
(3) \textit{Efficient expert utilization enables balanced trade-offs.} Unlike static models, TimeExpert dynamically activates the most relevant experts per input, resulting in stronger generalization to unseen datasets while keeping computation manageable.

\noindent\textbf{Performance Comparison on ActivityNet Captions.}
In \Cref{tab:anet_results}, we evaluate TimeExpert on the ActivityNet Captions \cite{krishna2017dense} dataset.
Notably, TimeExpert achieves the best performance across both Moment Retrieval and Dense Video Captioning tasks.
It surpasses TRACE, which was previously the strongest baseline, by improving CIDEr (+2.5) and F1 Score (+1.2) in Dense Video Captioning, while also significantly boosting $\text{R@}1_{\text{IoU}=0.5}$ (+1.5) and mIoU (+2.5) in Moment Retrieval.
These results highlight TimeExpert's ability to effectively model long-range temporal dependencies while efficiently leveraging a MoE framework for adaptive token routing.
Unlike previous methods, which rely on static expert activation, our model dynamically selects experts based on token-wise information, leading to improved alignment between textual queries and visual events.

\subsection{Ablation Studies}

\noindent\textbf{Design Choices.} As shown in the upper part of \Cref{tab:ablation_vtg_moe}, the removal of token-adaptive routing leads to a notable drop in all metrics, confirming its efficacy in efficiently allocating task tokens.
Moreover, removing separate encoders and decoding heads for causal event modeling completely disrupts the model's ability to follow instructions, reinforcing the necessity of structured task processing, and removing the auxiliary loss results in a slight performance drop across all tasks, particularly in Video Highlight Detection and Moment Retrieval. This indicates our proposed task-dependent auxiliary loss improves expert specialization by forcing task-aware routing, leading to better alignment between task tokens and relevant experts.

\noindent\textbf{Number of Input Frames.} As shown in the middle of \Cref{tab:ablation_vtg_moe}, increasing the number of frames generally improves performance across all tasks, as expected.
Notably, the transition from 64 to 128 frames boosts performance in Moment Retrieval and Video Highlight Detection, indicating that a larger temporal context helps refine event understanding.

\noindent\textbf{Number of Activated Experts.} As shown in the bottom part of \Cref{tab:ablation_vtg_moe}, there is a clear trend where the predefined number of activated experts directly influences performance.
Activating more experts improves performance, particularly in Video Highlight Detection (mAP +0.4 from 2 to 6 experts).
Nonetheless, beyond 6 experts, performance saturates, showing that excessive expert activation may introduce redundancy without additional benefits.

\noindent\textbf{Training Data.} We also conduct a data variant study, training our model on the same data recipe as TRACE to enable a direct comparison.
As shown in the bottom line of \Cref{tab:video_captioning}, TimeExpert still outperforms TRACE across all VTG tasks.
Compared to our full-data model, this variant exhibits slight fluctuations, with some metrics even improving (e.g., CIDEr and $\text{R@}1_{\text{IoU}=0.7}$), suggesting that TimeExpert generalizes well even with limited data.
These findings highlight TimeExpert's efficiency and adaptability, proving that our model design inherently enhances VTG performance, independent of dataset scale.

%% file: sec/5_conclusion.tex
\section{Conclusion}
\label{sec:conclusion}

In this work, we identify two critical limitations in existing VTG-specific Video-LLMs: inappropriate temporal event modeling and uniform task token processing. To address this, we propose \textsl{TimeExpert}, a novel framework that addresses these issues through a Mixture-of-Experts decoder that dynamically routes different task tokens (e.g., timestamps, saliency scores, and textual captions) to specialized experts, ensuring adaptive learning and mitigating task interference. Comprehensive experiments demonstrate TimeExpert's effectiveness across multiple VTG subtasks and general video understanding benchmarks, consistently outperforming existing VTG-specific Video-LLMs.